\documentclass[runningheads,a4paper]{llncs}

\usepackage{amssymb}
\usepackage{graphicx}
\usepackage{graphicx}
\usepackage{amsfonts}
\usepackage{amsmath}
\usepackage[table,dvipsnames]{xcolor}
\newcolumntype{L}{@{}>{\kern\tabcolsep}l<{\kern\tabcolsep}}
\usepackage{booktabs}
\usepackage{multirow}
\usepackage{enumitem}
\usepackage[misc]{ifsym}
\usepackage{algorithm,algorithmic}
\usepackage[table]{xcolor}
\newcolumntype{L}{@{}>{\kern\tabcolsep}l<{\kern\tabcolsep}}

\usepackage{xpatch,letltxmacro}

\usepackage{url}
\urldef{\mailsa}\path|{******, *******, *******, *******,|
\urldef{\mailsb}\path|********, *******, *******, *******,|
\urldef{\mailsc}\path|********, *********, *****}*******|    
\newcommand{\keywords}[1]{\par\addvspace\baselineskip
\noindent\keywordname\enspace\ignorespaces#1}

\usepackage[pages=some,placement=top]{background}

\DeclareMathOperator*{\minimum}{min}

\makeatletter

\newcommand{\ALC@it@nostep}{\item[]}
\LetLtxMacro\oldalgorithmic\algorithmic
\LetLtxMacro\endoldalgorithmic\endalgorithmic
\renewenvironment{algorithmic}[1][0]{%
  \oldalgorithmic[#1]%
  \xpatchcmd{\STATE}{\ALC@it}{\ALC@it@nostep}{}{}%
  \xpatchcmd{\FOR}{\ALC@it}{\ALC@it@nostep \textbf{\textit{Step 1 -- }}}{}{}%
  \xpatchcmd{\RETURN}{\ALC@it}{\ALC@it@nostep}{}{}%
  \xpatchcmd{\ENDFOR}{\ALC@it}{\ALC@it@nostep}{}{}%

}{\endoldalgorithmic}
\makeatother

\begin{document}

\backgroundsetup{contents=\underline{  To be published in the proceedings of Information Processing in Medical Imaging (IPMI) 2017  },color=black!100,opacity=1,scale=1.25,position={6.5,1.5}}
\BgThispage

\mainmatter  

\title{Risk Stratification of Lung Nodules Using 3D CNN-Based Multi-task Learning}

\titlerunning{Risk Stratification of Lung nodules}

%
%
\author{Sarfaraz Hussein\inst{1}
\and Kunlin Cao\inst{2}
\and Qi Song\inst{2}
\and Ulas Bagci\inst{1}
}

\authorrunning{S. Hussein et al.}

\institute{Center for Research in Computer Vision (CRCV) at University of Central Florida, Orlando, FL.\\
\email{shussein@knights.ucf.edu} , \email{bagci@crcv.ucf.edu}
\and CuraCloud Corporation, Seattle, WA.
}

\toctitle{Lecture Notes in Computer Science}
\tocauthor{Authors' Instructions}
\maketitle

\begin{abstract}
Risk stratification of lung nodules is a task of primary importance in lung cancer diagnosis. Any improvement in robust and accurate \emph{nodule characterization} can assist in identifying cancer stage, prognosis, and improving treatment planning. In this study, we propose a 3D Convolutional Neural Network (CNN) based nodule characterization strategy. With a completely 3D approach, we utilize the volumetric information from a CT scan which would be otherwise lost in the conventional 2D CNN based approaches. In order to address the need for a large amount for training data for CNN, we resort to \emph{transfer learning} to obtain highly discriminative features. Moreover, we also acquire the task dependent feature representation for six high-level nodule attributes and fuse this complementary information via a Multi-task learning (MTL) framework. Finally, we propose to incorporate potential disagreement among radiologists while scoring different nodule attributes in a graph regularized sparse multi-task learning. We evaluated our proposed approach on one of the largest publicly available lung nodule datasets comprising 1018 scans and obtained state-of-the-art results in regressing the malignancy scores.
\end{abstract}

\section{Introduction}

Cancer is the number-one cause of deaths in the world. Out of 8.2 million deaths due to cancer worldwide, lung cancer accounts for the highest number of mortalities i.e. 1.59 million \cite{stewart2016world}. Risk stratification of lung nodules can aid in identifying cancer stage leading to improved treatment and higher chances of survival. In addition, any significant development to accurately and automatically characterize lung nodules can save significant manual exertion as well as valuable time. 

Early diagnosis is one of the ways to reduce deaths related to lung cancer~\cite{van2015lung}. In this regard, lung screening programs are especially beneficial. Low Dose Computed Tomography (CT) scans are usually used to perform lung nodule diagnosis, including both detection and risk stratification. Although CT imaging remains the gold standard for lung cancer detection and diagnosis, Computer-Aided Diagnosis (CAD) and quantification tools are often necessary. Moreover, research in developing CAD algorithms can help explore the domain of imaging features and biomarkers which can be then studied by radiologists to further improve clinical decision making.

The development of a fast, robust and accurate system to perform risk stratification of lung nodules is therefore of significant importance. Specially the availability of large publicly available datasets such as LIDC-IDRI from Lung Image Database Consortium \cite{armato2011lung} has helped accelerate the research in this regard. However, the variability in nodule characteristics, including shape, size, intensity, location and uncertainty among radiologists' interpretation have made this problem particularly challenging. The advancement in machine learning methods, including the development of novel classification and feature learning techniques, has increased the efficacy of this task. However, there remains a substantial progress to be done in order to develop a CAD system attractive enough to be used in routine clinical evaluations of lung nodules.

\begin{figure*}[t]
\hspace{-1.0cm}
\includegraphics[width=140mm]{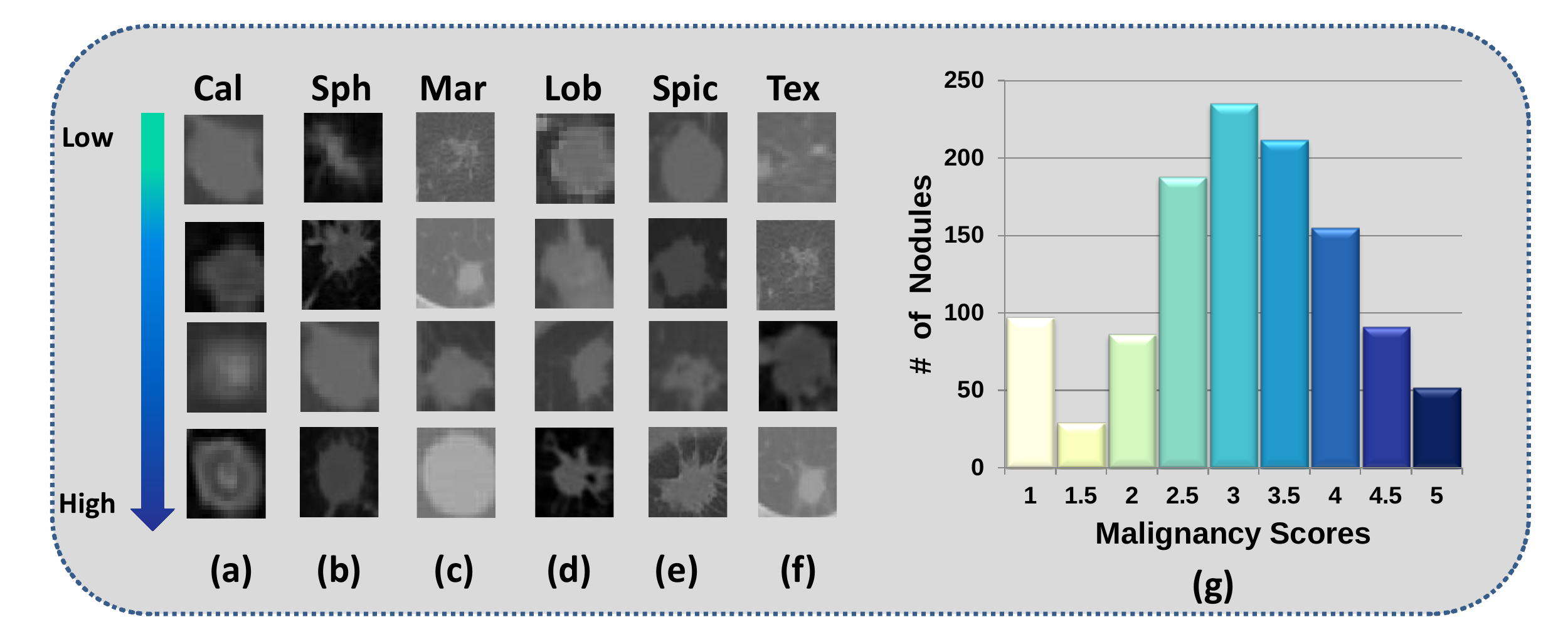}
\caption{Lung nodule attributes with different scores. As we move from the top (attribute missing) to the bottom (attribute with the highest prominence), the prominence of the attributes increases. Column (a) and (b) show calcified and spherical nodules; (c) represents margin where the top row is for poorly defined nodules and the bottom row shows well-defined nodules. Column (d) and (e) show lobulated and spiculated nodules whereas (f) represents nodules with different textures. The top row in (f) represents non-solid nodule and the bottom row shows solid nodule. The graph in (g) shows the number of nodules with different malignancy scores}
\label{fig:attributes}
\end{figure*}

\begin{figure*}[t]
\hspace{-0.6in}
\includegraphics[width=150mm,height=60mm]{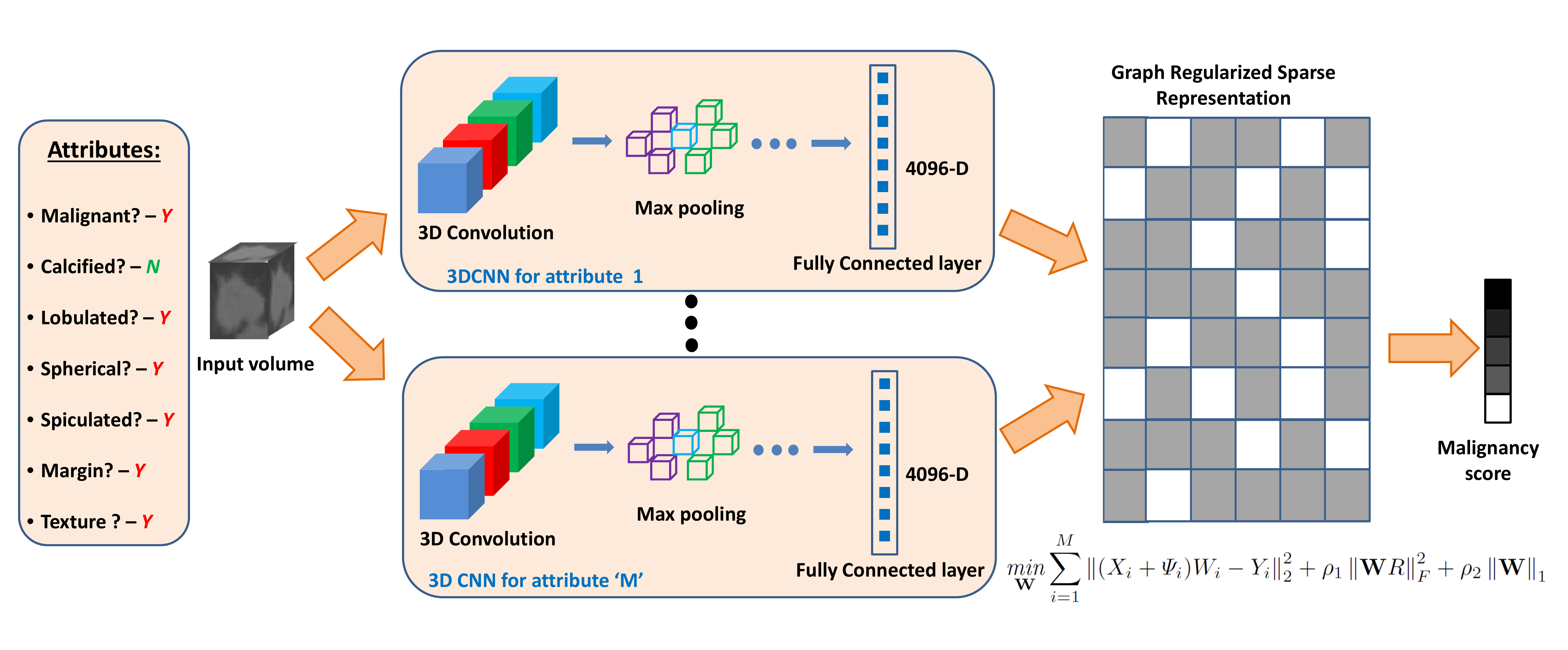}
\caption{An overview of the proposed approach. First, we fine-tune 3D CNNs using labels for malignancy and six attributes. Given the input volume, we pass it through different 3D CNNs each corresponding to an attribute (task). The network comprises 5 convolution, 5 max pooling, and 2 fully connected layers. We use the output from the first fully connected layer as the feature representation. The features from different CNNs are fused together using graph regularized sparse least square optimization function to obtain coefficient vectors corresponding to each task. During the testing phase, we multiply the feature representation of the testing image with the coefficient vector to obtain the malignancy score.}
\label{fig:workflow}
\end{figure*}

In this work, we address the challenge of risk-stratification of lung nodules in low-dose CT scans. Capitalizing on the significant progress of deep learning technologies for image classification and their potential applications in radiology~\cite{shin2016deep}, we propose a 3D Convolutional Neural Network (CNN) based approach for rich feature representation of lung nodules. We argue that the use of 3D CNN is paramount in the classification of lung nodules in low-dose CT scans which are 3D by nature. By using the conventional CNN methods, however, we implicitly lose the important volumetric information which can be very significant for accurate risk stratification. The superior performance of 3D CNN over 2D networks is well studied in \cite{tran2015learning}. We also avoid hand-crafted feature extraction, painstaking feature engineering, and parameter tuning. Moreover, any information about six high-level nodule attributes such as calcification, sphericity, margin, lobulation, spiculation and texture (Figure~\ref{fig:attributes}) can help in improving the benign-malignant risk assessment of the nodules. Taking forward this idea, we identify features corresponding to these high-level nodule attributes and fuse them in a multi-task learning framework to obtain the final risk assessment scores. An overview of the proposed approach is presented in Figure~\ref{fig:workflow}. Overall, our main contributions in this work can be summarized as follows:
\begin{itemize}[topsep=0em,leftmargin=*]
\itemsep0.7em 
\item We propose a 3D CNN based method to utilize the volumetric information from a CT scan which would be otherwise lost in the conventional 2D CNN based approaches. Moreover, we also circumvent the need for a large amount of volumetric training data to train the 3D network by transfer learning. We use the CT data to fine-tune a network which is trained on 1 million videos. To the best of our knowledge, our work is the first to empirically validate the success of transfer learning of a 3D network for lung nodules. 
\item We perform experimental evaluations on one of the largest publicly available datasets comprising lung nodules from more than 1000 low-dose CT scans.
\item We employ graph regularized sparse multi-task learning to fuse the complementary feature information from high-level nodule attributes for malignancy determination. We also propose a scoring function to measure the inconsistency in risk assessment among different experts (radiologists). 
\end{itemize}

\section{Related Work}
Conventionally, the characterization of lung nodules comprised nodule segmentation, extraction of hand-crafted imaging features, followed by the application of an off-the-shelf classifier/regressor. The method by Uchiyama et al. \cite{uchiyama2003quantitative} was based on the extraction of various physical measures, including intensity statistics and then classification using Artificial Neural Networks. El-Baz et al. \cite{el20113d} first segmented the lung nodules using appearance-based models and used spherical harmonic analysis to perform shape analysis. The final step was the classification using $k$-nearest neighbor. Proposing a study based on texture analysis, Han et al. \cite{han2015texture} extracted 2D texture features such as Haralick, Gabor and Local Binary Patterns (LBP) and extended them to 3D. Support Vector Machine (SVM) was employed to perform the classification. In another classical work by Way et al. \cite{way2006computer}, segmentation is performed using 3D active contours followed by the extraction of texture features from the rubber band straightening transform of the surrounding voxels. The classification was performed using Linear Discriminant Analysis (LDA) classifier. In another study, Lee et al. \cite{lee2010computer} proposed a feature selection based approach using both imaging and clinical data. An ensemble classifier, combining genetic algorithm (GA) and random subspace method (RSM) was then used to gauge feature relevance and information content. Finally, LDA was employed to perform classification on the reduced feature set.

Following up on the success of deep learning, the medical imaging community has moved from feature engineering to feature learning. In those frameworks, CNN had been used for feature extraction and an off-the-shelf classifier such as Random Forest (RF) was employed for classification \cite{kumar2015lung,buty2016characterization}. Recently, Buty et al. \cite{buty2016characterization} combined spherical harmonics along with deep CNN features and then classified them using RF. However, the use of CNN for lung nodule classification has been confined to 2D image analysis~\cite{chen2016bridging}, thus falling short of utilizing the important volumetric and contextual information. 

Moreover, the use of high-level image attributes had been found to be instrumental in the risk assessment and classification of lung nodules. In an effort to study the relationship between nodules attributes and malignancy, Furuya et al. \cite{furuya1999new} found that in a particular dataset, 82\% of the lobulated, 97\% of the densely spiculated, 93\% of the ragged and 100\% of the halo nodules were malignant. Moreover, 66\% of the round nodules were found to be benign. Inspired by this study, in this work we utilize 3D CNN to learn discriminative feature set corresponding to each of the 6 attributes. We then fuse these feature representations via MTL to determine the malignancy likelihood.  
\section{Method}
\subsection{Problem Formulation}
Let $X=[x_1,x_2 \dots x_n]\in\mathbb{R}^{n \times d}$ be the data matrix comprising features from $n$ data points in $\mathbb{R}^d$. Each sample corresponds with a regression score given by $Y=[y_1,y_2 \dots y_n]$ where $Y\in \mathbb{R}^{n \times 1}$. Here the objective is to learn the coefficient vector or the regression estimator $W$ from the training data. In this case, the $\ell_1$ regularized least square regressor is defined as: 

\begin{equation}
\minimum_{W}\left \|XW-Y  \right \|_{2}^{2}+\lambda\left \| W \right \|_{1},
\label{eq:l1norm}
\end{equation}

\noindent where $\lambda$ controls the sparsity level for coefficient vector $W=[w_1,w_2 \dots w_d]$. The problem in Eq.~\ref{eq:l1norm} is an \emph{unconstrained convex optimization} problem, and it remains non-differentiable when $w_i=0$. Hence, the closed form solution corresponding to the global minimum for Eq.~\ref{eq:l1norm} is not possible. Thus, the above equation is represented in the following way as a \emph{constrained optimization} function:

\begin{equation}
\begin{split}
\minimum_{W}  \left \|XW-Y  \right \|_{2}^{2},\\
s.t.   \left \| W \right \|_1 \leq t,
\end{split}
\label{eq:l1normc}
\end{equation}

\noindent where $t$ is inversely proportional to $\lambda$. In the representation given in Eq.~\ref{eq:l1normc}, both optimization function and the constraint are convex. 

\subsection{Network Architecture and Transfer Learning}
We use the lung nodules dataset to fine-tune a 3D CNN trained on Sports-1M dataset \cite{karpathy2014large}. The sports dataset comprises 1 million videos with 487 classes. In the absence of a large number of training examples from lung nodules, we use transfer learning strategy to obtain rich feature representation from a larger dataset (Sports-1M) for lung nodule characterization. The Sports-1M dataset is used to train a 3D CNN~\cite{tran2015learning}. The network comprises 5 convolution, 5 max-pooling, 2 fully-connected and 1 soft-max classification layers. The input to the network is 3$\times$16$\times$128$\times$171 where there are 16 non-overlapping slices in the input volume. The first 2 convolution layers have 64 and 128 filters respectively, whereas there are 256 filters in the last 3 layers. The outputs of the fully connected layers are of 4096 dimensions.
\subsection{Multi-task learning}
Consider a problem with $M$ tasks representing different attributes corresponding to a given dataset $D$.  These tasks may be related and share some feature representation, both of which are unknown. The goal in Multi-task learning (MTL) is to perform joint learning of these tasks while exploiting dependencies in feature space so as to improve regressing one task using the others. In contrast to multi-label learning, tasks may have different features in MTL. Each task has model parameters denoted by $W_m$, used to regress the corresponding task $m$. Moreover, when $\textbf{W}$ $=[W_1,W_2 \dots W_M] \in\mathbb{R}^{M \times d}$ represents a rectangular matrix, rank is considered as an extension to the cardinality. In that case, trace norm, which is the sum of singular values is a replacement to the $\ell_1$-norm. Trace norm, also known as nuclear norm is the convex envelope of the rank of a matrix (which is non-convex), where the matrices are considered on a unit ball. By replacing $\ell_1$-norm with trace norm in Eq.~\ref{eq:l1norm}, the trace norm regularized least square loss function is given by:

\begin{equation}
\minimum_{\mathbf{W}}\sum_{i=1}^{M} \left \|X_iW_i-Y_i  \right \|_{2}^{2}+\rho\left \| \mathbf{W} \right \|_{*},
\label{eq:mtltrace}
\end{equation}

\noindent where $\rho$ tunes the rank of the matrix \textbf{W}, and trace-norm is defined as: $\left \| \mathbf{W} \right \|_{*}=\sum_{i=1}\sigma_i(\mathbf{W})$ with $\sigma$ representing singular values.

Another regularizer, pertinent to MTL, is the regularization on the graph representing the relationship between the tasks \cite{evgeniou2004regularized,zhou2011malsar}. Consider a complete graph $G=(V,\mathcal{E})$, such that nodes $V$ represent the tasks and the edges $\mathcal{E}$ encode any relativity between the tasks. The complete graph can be represented as a structure matrix $S=[e^1,e^2 \dots e^{\left \| \mathcal{E} \right \|}]$ and the difference between all the pairs connected in the graph is penalized by the following regularizer:

\begin{equation}
\left \|\mathbf{W}S  \right \|_{F}^{2}=\sum_{i=1}^{\left \| \mathcal{E} \right \|} \left \|\mathbf{W}e^i \right \|_{2}^{2}=
\sum_{i=1}^{\left \| \mathcal{E} \right \|} \left \|\mathbf{W}_{e^{i}_{a}}-\mathbf{W}_{e^{i}_{b}} \right \|_{2}^{2}.
\label{eq:mtltrace}
\end{equation}

Herein, $e^{i}_{a}$, $e^{i}_{b}$ are the edges between the nodes $a$ and $b$. The above regularizer can also be written as:

\begin{equation}
\left \|\mathbf{W}S  \right \|_{F}^{2}=\text{tr}((\mathbf{W}S)^T (\mathbf{W}S))=\text{tr}(\mathbf{W}SS^T \mathbf{W}^T)=\text{tr}(\mathbf{W} \mathcal{L}\mathbf{W}^T),
\label{eq:mtltrace}
\end{equation}

\noindent where `tr' represents trace of a matrix and $\mathcal{L}=SS^T$ is the Laplacian matrix. Since there may exist disagreements between the scores from different experts (radiologists), we propose a scoring function to measure potential inconsistencies:

\begin{equation}
\mbox{\large\(\Psi(j) =(e^\frac{-\sum_{i}(x_i^j-\mu^j )^{2}}{2\sigma^p} )^{-1}\)}.
\end{equation}

The inconsistency measure corresponding to a particular example $j$ is represented by $\Psi(j)$. $x_i^j$ is the score given by the expert (radiologist) $i$ and $\mu^j$ and $\sigma^j$ denote mean and standard deviation of the scores, respectively. Here, for simplicity, we have dropped the index for the task; however, note that the inconsistency measure is computed for all the tasks. The final proposed graph regularized sparse least square optimization function with the inconsistency measure can then be written as:

\begin{equation}
\minimum_{\mathbf{W}}\sum_{i=1}^{M}\overset{{\textcircled{1}}}{\overbrace{\left \|(X_i+\Psi_i)W_i-Y_i  \right \|_{2}^{2}}}+\overset{{\textcircled{2}}}{\overbrace{\rho_1\left \| \mathbf{W}S \right \|_{F}^{2}}}+\overset{{\textcircled{3}}}{\overbrace{\rho_2\left \| \mathbf{W} \right \|_{1}}},
\label{eq:finaleq}
\end{equation}


\noindent where $\rho_1$ controls the level of penalty for graph structure and $\rho_2$ controls the sparsity. In the above optimization, the least square loss function \textcircled{1} considers tasks to be decoupled whereas \textcircled{2} and \textcircled{3} consider the interdependencies between different tasks.

\subsection{Optimization}
The optimization function in Eq. \ref{eq:finaleq} cannot be solved through standard gradient descent because the $\ell_1-$norm is not differentiable at $\mathbf{W}=0$. Since the optimization function in Eq.~\ref{eq:finaleq} has both smooth and non-smooth convex parts, estimating the non-smooth part can help solve the optimization function. Therefore, \emph{accelerated proximal gradient method} \cite{nesterov2013introductory,parikh2014proximal} is employed to solve the Eq. \ref{eq:finaleq}. The accelerated proximal method is the first order gradient method with a complexity of $O(1/k^2)$, where $k$ is the iteration counter. Note that in Eq.~\ref{eq:finaleq}, the $\ell_1$-norm comprises the non-smooth part and the proximal operator is used for its estimation. The steps in the proposed approach are summarized in Algorithm 1. 

 \begin{algorithm*}[t]
 \caption{Algorithm for the proposed MTL method}
 \begin{algorithmic}[1]
 \renewcommand{\algorithmicrequire}{\textbf{Input:}}
 \renewcommand{\algorithmicensure}{\textbf{Output:}}
 \REQUIRE Generated features from 3D CNN: $X_M^N$ for $M$ attributes and $N$ examples\\
         \qquad Attributes scores: $Y_M^N$
 \ENSURE  Coefficient matrix $\mathbf{W}$
  \STATE \textbf{\textit{Step 1 --}} \textbf{for} each task $i = 1$ to $M$ and each example $j = 1$ to $N$ \textbf{do}\\
  \qquad \qquad \qquad Solve equation (6) to find $\Psi$\\
  \quad \quad \qquad \textbf{end for}
  \STATE \textbf{\textit{Step 2 --}} Formulate objective function as in equation (7)
  \STATE \textbf{\textit{Step 3 --}} Use accelerated proximal gradient method to optimize equation (7)\\
 \RETURN $\mathbf{W}$ 
 \end{algorithmic} 
 \end{algorithm*}

\section{Experiments}

\subsection{Data}
For evaluating our proposed approach, we used LIDC-IDRI dataset from Lung Image Database Consortium \cite{armato2011lung}, which is one of the largest publicly available lung cancer screening datasets. There were 1018 CT scans in the dataset, where the slice thickness varied from 0.45 mm to 5.0 mm. The nodules having diameters larger than or equal to 3 mm were annotated by at most four radiologists.

The nodules which were annotated by at least three radiologists were used for the evaluations. There were 1340 nodules satisfying this criterion. We used the mean malignancy and attribute scores of different radiologists for experiments. The nodules have ratings corresponding to malignancy and the other six attributes which are (i) calcification, (ii) lobulation, (iii) spiculation, (iv) sphericity, (v) margin and (vi) texture. The malignancy ratings varied from 1 to 5 where 1 indicated benign and 5 represented highly malignant nodules. We excluded nodules with an average score equal to 3 to account for the indecision among the radiologists. Our final dataset consisted of 635 benign and 509 malignant nodules for classification. The images were resampled to have 0.5 mm spacing in each dimension.
\subsection{Results}
We used the 3D CNN trained on Sports-1M dataset \cite{karpathy2014large} which had 487 classes. We fine-tuned the network using samples from lung nodule dataset. In order to generate the binary labels for the six attributes and the malignancy, we used the center point and gave positive (or negative) labels to samples having scores greater (or lesser) than the center point. In the context of our work, tasks represented six attributes and malignancy. We fine-tuned the network with these 7 tasks and performed 10 fold cross validation. By fine-tuning the network, we circumvented the need to have a large amount of training data. Since the 3D network was trained on image sequences with 3 channels and with at least 16 frames, we replicated the same gray level axial channel for the other two.
\begin{table*}
\begin{center}
\caption{Classification accuracy and mean absolute score difference of the proposed multi-task learning method in comparison with the other methods.}
\label{table:Results}
\label{table:quan_acc}
\begin{tabular}{l@{\hspace{0.7in}}c@{\hspace{0.4in}}c@{\hspace{0.2in}}c}
\toprule[1.5pt] \multirow{2}{*}{\textbf{Methods}}   & \multirow{2}{*}{\textbf{Accuracy}} & \multirow{2}{*}{\textbf{Mean score diff.}} \\ 
\\
\cmidrule(r){1-4}
GIST features with LASSO    &      76.83\%     &      0.6753   &        \\
3D CNN MTL with Trace norm &      80.08\%     &      0.6259  &        \\
\textbf{Proposed method (Equation 7)}     &     \textbf{91.26\%}    &      \textbf{0.4593}   &        \\
\toprule[1.5pt]
\end{tabular}
\end{center}
\end{table*}
Moreover, we also ensured that all input volumes have 16 slices by interpolation when necessary. We used the 4096-dimensional output from the first fully connected layer of the 3D CNN as a feature representation.

\begin{figure*}[t]
\hspace{-2.25cm}
\includegraphics[width=165mm,height=55mm]{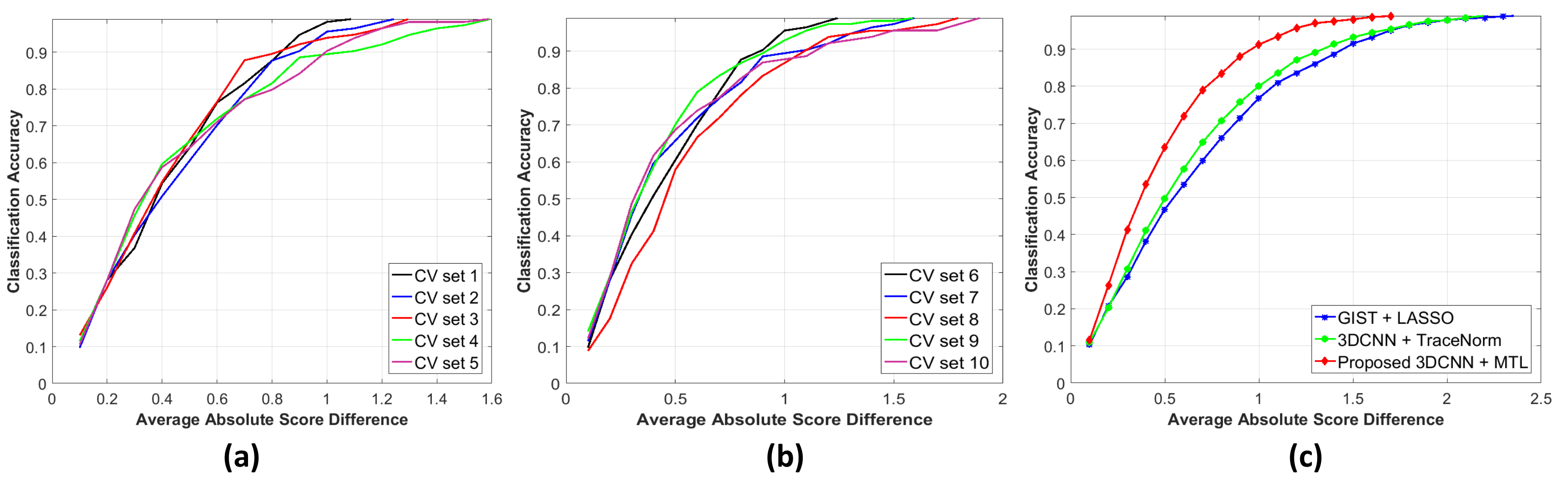}
\caption{Plots to show classification accuracy against various threshold values for average absolute score difference. The graphs in (a) and (b) represent results from 10 different cross validation (CV) sets. It can be seen that the classification accuracy increases, as we increase the threshold value for absolute score difference. The graph (c) shows the improved performance of the proposed method in comparison with GIST+LASSO and 3DCNN with Trace Norm.}
\label{fig:graph}
\end{figure*}

To find the structure matrix $S$, we computed the correlation between tasks by finding an initial normalized coefficient matrix $\mathbf{W}$ using lasso with least square loss function and followed by computing the correlation coefficient matrix \cite{zhou2011malsar}. We then apply a threshold on the correlation coefficient matrix to obtain a binary graph structure matrix. For testing, we multiply the features from network trained on malignancy with the corresponding coefficient vector $W$ to obtain the score.
 
For evaluation, we used metrics for both classification and regression. We calculated classification accuracy by considering classification to be successful if the predicted score lies in $\pm$1 of the true score. We also reported average absolute score difference between the predicted score and the true score. Table \ref{table:quan_acc} shows the comparison of our proposed Multi-task learning method with GIST features~\cite{GIST} +LASSO and 3D CNN Multi-task learning with trace norm. Our proposed graph regularized MTL outperforms the other methods with a significant margin. Our approach improves the classification accuracy over GIST features by about 15\% and over trace norm regularization by 11\%. Moreover, the average absolute score difference reduces by 32\% and 27\% when compared with GIST and trace norm respectively.

\begin{figure*}[t]
\hspace{-0.5cm}
\includegraphics[width=130mm]{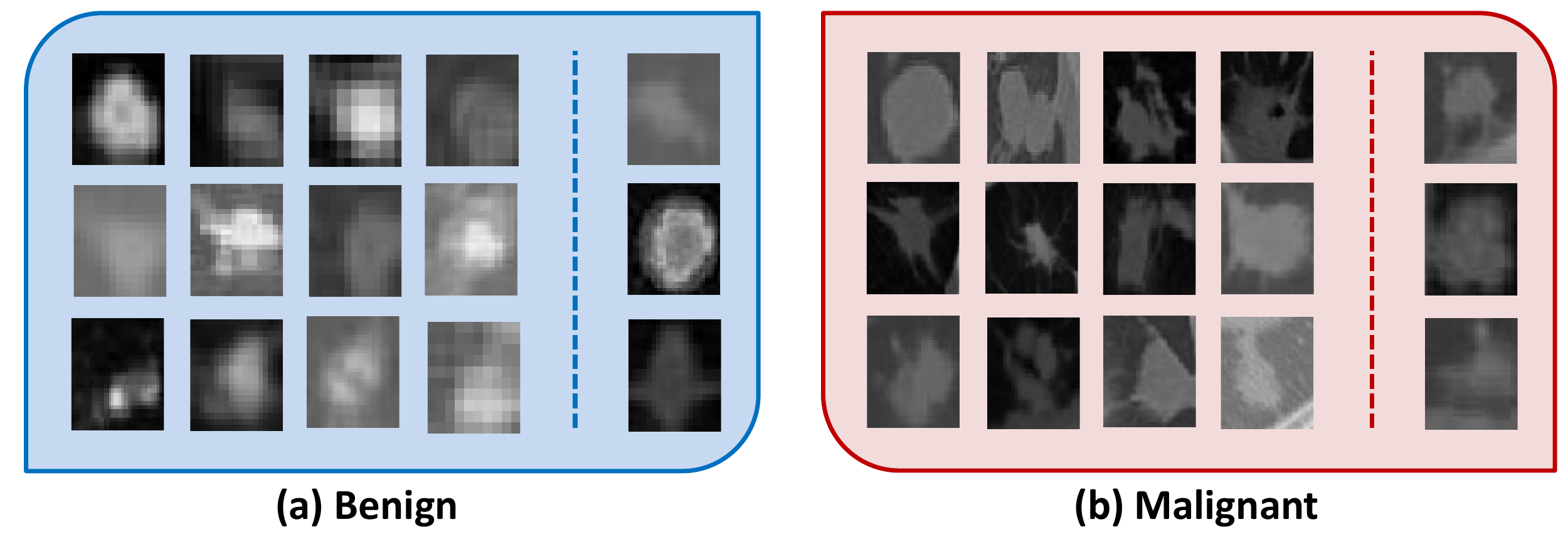}
\caption{Qualitative results using our proposed approach. (a) and (b) show axial views of benign and malignant nodules respectively, where first three columns consist of successful cases (where prediction was within $\pm$1 of the expert score) and the last column (after dotted line) shows failure cases.}
\label{fig:qual}
\end{figure*}

We also plotted classification accuracy against different thresholds for average absolute score difference. Figure \ref{fig:graph} (a) and (b) show the plot on different cross-validation sets. It can be noticed that across different validation sets, the predicted malignancy scores of around 70\% of the nodules lie within a margin of $\pm$0.6 which increases to around 90\% when $\pm$1 margin is used. Figure   \ref{fig:graph} (c) shows the comparison with the other methods, where the proposed approach outperforms them over all values of average absolute score difference. Figure~\ref{fig:qual} shows the qualitative results from our proposed approach.

\begin{figure*}[t]
\hspace{-2cm}
\includegraphics[width=160mm,height=70mm]{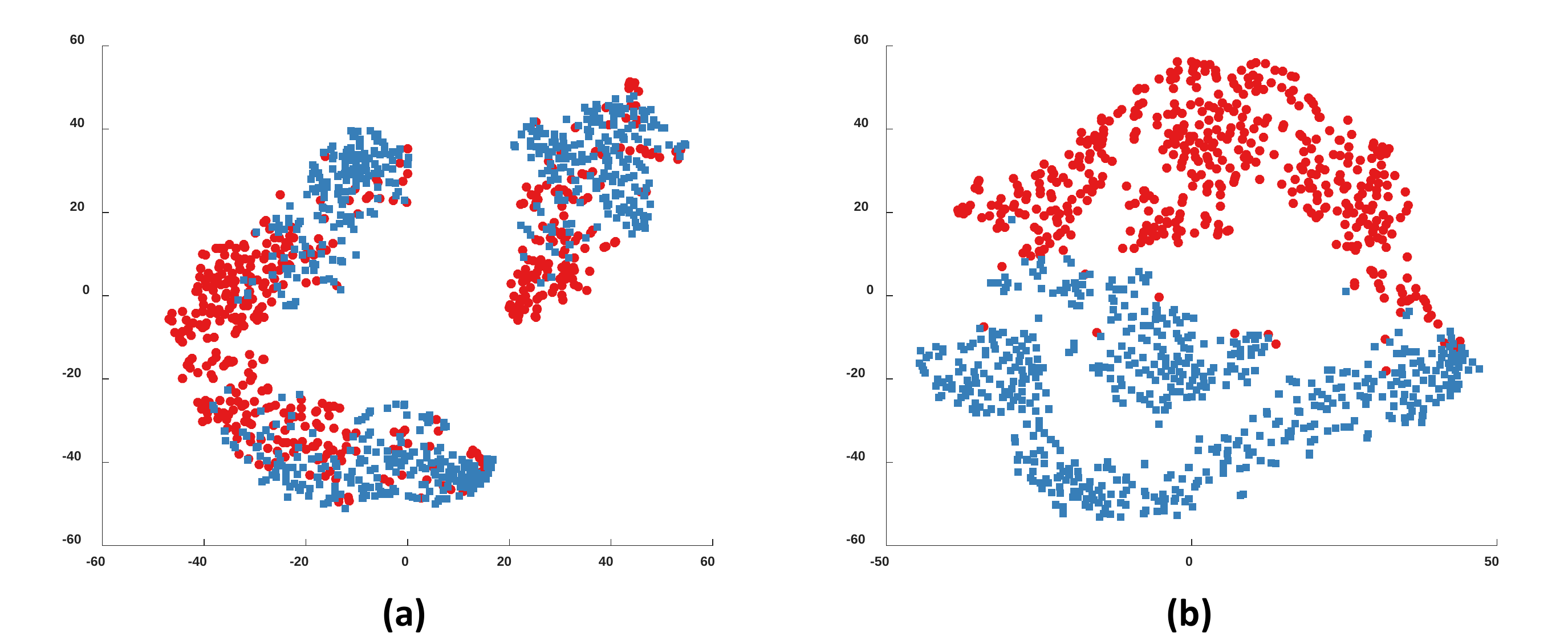}
\caption{Effect of fine-tuning on 3D CNN features. t-SNE visualization for features obtained from (a) pre-trained network and (b) network after fine-tuning. Separation between features belonging to two classes, i.e. benign nodules (represented in blue) and malignant nodules (shown in red) can be readily observed in (b).}
\label{fig:tsne}
\end{figure*}

In order to evaluate the significance of transfer learning via fine-tuning, we project the features onto a low dimensional space. This is done by computing the proximity, between boundary points using t-distributed stochastic neighborhood embedding (t-SNE)~\cite{tsne}. As our feature space is high dimensional (4096-dimension), t-SNE is useful in revealing the structure of data at different scales. It can be seen in Figure~\ref{fig:tsne} that fine-tuning the network on the lung nodule dataset distinctively improves the separation between benign and malignant classes.

\section{Discussion and Conclusion}
In this work, we proposed a framework to stratify the malignancy of lung nodules using 3D CNN and graph regularized sparse multi-task learning. To the best of our knowledge, this is for the first time, transfer learning is employed over 3D CNN to improve lung nodule characterization. The task of data collection, especially in medical imaging fields, is highly regulated and the availability of experts for annotating these images is restricted. In this scenario, leveraging on the availability of crowdsourced and annotated data such as user captured videos can be instrumental in training discriminative models. However, given the diversity in data from these two domains (i.e. medical and non-medical user collected videos), it is vital to perform transfer learning from source domain (user collected videos) to the target domain (medical imaging data). To establish this observation and to visualize features, we used t-SNE to project high dimensional features onto a low dimensional space (2D space), where the separation between classes was evident in the case of transfer learning.

Moreover, in this work, we also empirically explored the importance of high-level nodule attributes such as calcification, sphericity, lobulation and others to improve malignancy determination. Rather than manually determining these attributes we used 3D CNN to learn discriminative features corresponding to these attributes. The 3D CNN based features from these attributes are fused in a graph regularized sparse multi-task learning.

Another important imaging modality for lung nodule diagnosis is Positron Emission Tomography (PET). It has been found that the combination of PET and CT can improve the diagnostic accuracy of solitary lung nodules~\cite{wang2014evidence}. With the increase in the availability of PET/CT scanners, our future work will involve their utilization for simultaneous detection and characterization of pulmonary nodules.

\bibliographystyle{splncs04}
\bibliography{egbib}

\end{document}